\documentclass[runningheads]{llncs}
\usepackage{graphicx}

\usepackage{tikz}
\usepackage{comment} 
\usepackage{amsmath,amssymb} 
\usepackage{color}

\usepackage{multirow}
\usepackage{pifont}
\usepackage{tabularx}
\usepackage{bm}
\usepackage{enumitem} 
\usepackage{hyperref}
\setlist{nolistsep}

\hypersetup{
	colorlinks=true,
	urlcolor=magenta,
}

\usepackage{pdfpages}

\newcolumntype{L}[1]{>{\raggedright\arraybackslash}p{#1}}
\newcolumntype{C}[1]{>{\centering\arraybackslash}p{#1}}
\newcolumntype{R}[1]{>{\raggedleft\arraybackslash}p{#1}}
\newcommand{\cmark}{\ding{51}}%
\newcommand{\xmark}{\ding{55}}%
\newcommand{\etal}{\textit{et al}.}
\newcommand{\ie}{\textit{i}.\textit{e}.,}

\begin{document}
\pagestyle{headings}
\mainmatter
\def\ECCVSubNumber{2019}  

\title{Learning Modality Interaction for\\
	Temporal Sentence Localization and\\ Event Captioning in Videos} 

\titlerunning{Learning Modality Interaction in Videos}
%
\author{Shaoxiang Chen\inst{1}\thanks{Part of the work is done when the author was an intern at Tencent AI Lab.}\and
Wenhao Jiang\inst{2} \and
Wei Liu\inst{2}\and
Yu-Gang Jiang\inst{1}\thanks{Corresponding author.}}
\authorrunning{S. Chen et al.}
%
\institute{Shanghai Key Lab of Intelligent Information Processing, \\School of Computer Science, Fudan University\\
\and
Tencent AI Lab\\
\email{\{sxchen13, ygj\}@fudan.edu.cn, cswhjiang@gmail.com, wl2223@columbia.edu}}
\maketitle

\begin{abstract}
	
Automatically generating sentences to describe events and temporally localizing sentences in a video are two important tasks that bridge language and videos. Recent techniques leverage the multimodal nature of videos by using off-the-shelf features to represent videos, but interactions between modalities are rarely explored. Inspired by the fact that there exist cross-modal interactions in the human brain, we propose a novel method for learning pairwise modality interactions in order to better exploit complementary information for each pair of modalities in videos and thus improve performances on both tasks. We model modality interaction in both the sequence and channel levels in a pairwise fashion, and the pairwise interaction also provides some explainability for the predictions of target tasks.
We demonstrate the effectiveness of our method and validate specific design choices through extensive ablation studies. Our method turns out to achieve state-of-the-art performances on four standard benchmark datasets: MSVD and MSR-VTT (event captioning task), and Charades-STA and ActivityNet Captions (temporal sentence localization task).

\keywords{Temporal Sentence Localization \and Event Captioning in Videos \and Modality Interaction}
\end{abstract}

\section{Introduction}

Neuroscience researches~\cite{Calvert2001,Baier2006,Eckert2008} have discovered that the early sensory processing chains in the human brain are not unimodal, information processing in one modality (\textit{e.g.}, auditory) can affect another (\textit{e.g.}, visual), and there is a system in the brain for modulating cross-modal interactions.
However, modality interactions are largely overlooked in the research of high-level video understanding tasks, such as event captioning~\cite{Yao2015,Venugopalan2015,Venugopalan2015a} and temporal sentence localization~\cite{Gao2017,Liu2018,Chen2018}.
Both tasks involve natural language descriptions and are substantially more challenging than recognition tasks. 
Thus, it is crucial to utilize information from each of the available modalities and capture inter-modality complementary information to better tackle these tasks. 

Recent event captioning methods~\cite{Pan2016,Song2017,Wang2018a,Chen2019a,Hu2019} mostly adopt an encoder-decoder structure, where the encoder aggregates video features and the decoder (usually LSTM~\cite{Hochreiter1997} or GRU~\cite{Chung2014}) generates sentences based on the aggregation results. 
The video features stem mainly from the visual appearance modality, which are usually extracted with off-the-shelf CNNs (Convolutional Neural Networks)~\cite{Szegedy2017,He2016,Szegedy2015,Simonyan2015} that are pre-trained to recognize objects and can output high-level visual representations for still images. 
Using features from the visual modality solely can generally work well on video event captioning. 
Recent works \cite{Yu2016,Pan2016,Chen2019a,Shi2019,Chen2017,Shen2017,Rahman2019} suggest that further improvements can be obtained by additionally leveraging motion and audio representations. 
However, the limitation of these works is that the features from multiple modalities are simply concatenated without considering their relative importances or the high-level interactions among them, so the great potential of multiple modalities has not been fully explored.
There exist a few works~\cite{Jin2019,Hori2017,Zhang2017,Xu2017} that learn to assign importance weights to individual modalities via cross-modal attention in the encoder, but modality interactions are still not explicitly handled. 
Temporal sentence localization in videos is a relatively new problem~\cite{Gao2017}. 
Although various approaches~\cite{Song2018,Yuan2019,Chen2019} have been proposed and significant progresses have been made, this problem has not been discussed in a multimodal setting. 
Most recently, Rahman~\etal~\cite{Rahman2019} emphasized the importance of jointly considering video and audio 
to tackle dense event captioning, in which sentence localization is a subtask. 
Apart from the visual, motion, and audio modalities, utilizing semantic attributes is gaining popularity in recent methods~\cite{Aafaq2019,Long2018,Chen2019,Wang2019} for both event captioning and sentence localization.

In order to better exploit multimodal features for understanding video contents, 
we propose a novel and generic method for modeling modality interactions that can be leveraged to effectively improve performances on both the sentence localization and event captioning tasks.  
Our proposed Pairwise Modality Interaction (PMI) explicitly models sequence-level interactions between each pair of feature sequences by using a channel-gated bilinear model, and the outputs of each interacting pair are fused with importance weights. Such a modeling provides some explainability for the predictions.

Our main contributions are as follows:
\begin{itemize}
	\item We propose a novel multimodal interaction method that uses a Channel-Gated Modality Interaction model to compute pairwise modality interactions (PMI), which better exploits intra- and inter-modality information in videos. Utilizing PMI achieves significant improvements on both the video event captioning and temporal sentence localization tasks.
	\item Based on modality interaction within video and text, we further propose a novel sentence localization method that builds video-text local interaction for better predicting the position-wise video-text relevance. To the best of our knowledge, this is also the first work that addresses sentence localization in a multimodal setting.
	\item Extensive experiments on the MSVD, MSR-VTT, ActivityNet Captions, and Charades-STA datasets verify the superiority of our method compared against state-of-the-art methods on both tasks. 
\end{itemize}

\section{Related Works}

\textbf{Temporal Sentence Localization} 
Gao~\etal~\cite{Gao2017} proposed the temporal sentence localization task recently, 
and it has attracted growing interests from both the computer vision and natural language processing communities. Approaches for this task can be roughly divided into two groups, \textit{i.e.}, proposal-based methods and proposal-free methods.  
TALL~\cite{Gao2017} uses a multimodal processing module to fuse visual and textual features for sliding window proposals, and then predicts a ranking score and temporal boundaries for each proposal. 
NSGV~\cite{Chen2018} performs interaction between sequentially encoded sentence and video via an LSTM, and then predicts $K$ proposals at each time step. 
Proposal-free methods usually regress the temporal boundaries. As the most representative one, ABLR~\cite{Yuan2019} iteratively applies  co-attention between visual and textual features to encourage interactions, and finally uses the interacted features to predict temporal boundaries. 

\textbf{Event Captioning} 
The S2VT~\cite{Venugopalan2015} method is the first attempt at solving video captioning using an encoder-decoder network, in which two layers of LSTMs~\cite{Hochreiter1997} first encode the CNN-extracted video features and then predict a sentence word-by-word.
Later works are mostly based on the encoder-decoder structure, and improvements are made for either the encoder or decoder. 
Yao~\etal~\cite{Yao2015} applied temporal attention to the video features, which enables the encoder to assign an importance weight to each video feature during decoding, and this method is also widely adopted by the following works. 
Some works~\cite{Pan2016,Baraldi2017,Chen2018a,Zhu2017,WangJW2018} tried to improve the encoder by considering the temporal structures inside videos. 
Another group of works~\cite{Yang2017,Li2017,Chen2019a} are focused on exploiting spatial information in video frames 
by applying a dynamic attention mechanism to aggregate frame features spatially.
Utilizing multimodal (appearance, motion, and audio) features is also common in recent works, but only a few works~\cite{Hori2017,Zhang2017,Xu2017,Long2018} tried to handle the relative importances among different modalities using cross-modal attention.
Most recently, some works~\cite{Aafaq2019,Zhang2019,Long2018} have proven that incorporating object/semantic attributes into video captioning is effective. 
As for the decoder, LSTM has been commonly used as the decoder for video captioning, and some recent attempts have also been made to using non-recurrent decoders such as CNN~\cite{Chen2019b} or the Transformer~\cite{Zhou2018} structure.

\textbf{Modality Interaction} 
There are some works trying to use self-attention to model modality interaction. 
Self-attention has been proven effective on both vision~\cite{Wang2018c} and language~\cite{Vaswani2017} tasks.
Its effectiveness in sequence modeling can be attributed to that it computes a response at one position by attending to all positions in a sequence, which better captures long-range dependencies. 
AutoInt~\cite{Song2019} concatenates features from different modalities and then feeds them to a multi-head self-attention module for capturing interactions. 
For the referred image segmentation task, Ye~\etal~\cite{Ye2019} introduced CMSA (Cross-Modal Self-Attention), which operates on the concatenation of visual features, word embeddings, and spatial coordinates to model long-range dependencies between words and spatial regions.
DFAF~\cite{Gao2019a} is a visual question answering (VQA) method, which applies self-attention for regional feature sequences and word embedding sequences to model inter-modality interactions, and also models intra-modality interactions for each sequence. 
We note that modality interaction is common in VQA methods, but they usually pool the multimodal feature sequences into a single vector using bilinear or multi-linear pooling~\cite{Fukui2016,Kim2018,Kim2017,Liu2018a}. And VQA methods are more focused on the interaction between visual and textual modalities, so they do not fully exploit the modality interactions within videos.

Compared to these existing methods, our proposed Pairwise Modality Interaction (PMI) has two distinctive features: 
(1) modality interactions are captured in a pairwise fashion, and information flow between each pair of modalities in videos is explicitly considered in both the sequence level and channel level;
(2) the interaction does not pool the feature sequences (\ie~temporal dimension is preserved), and the interaction results are fused by their importance weights to provide some explainability.

\section{Proposed Approach}

\begin{figure*}[t]
	\centering \includegraphics[width=.95\textwidth]{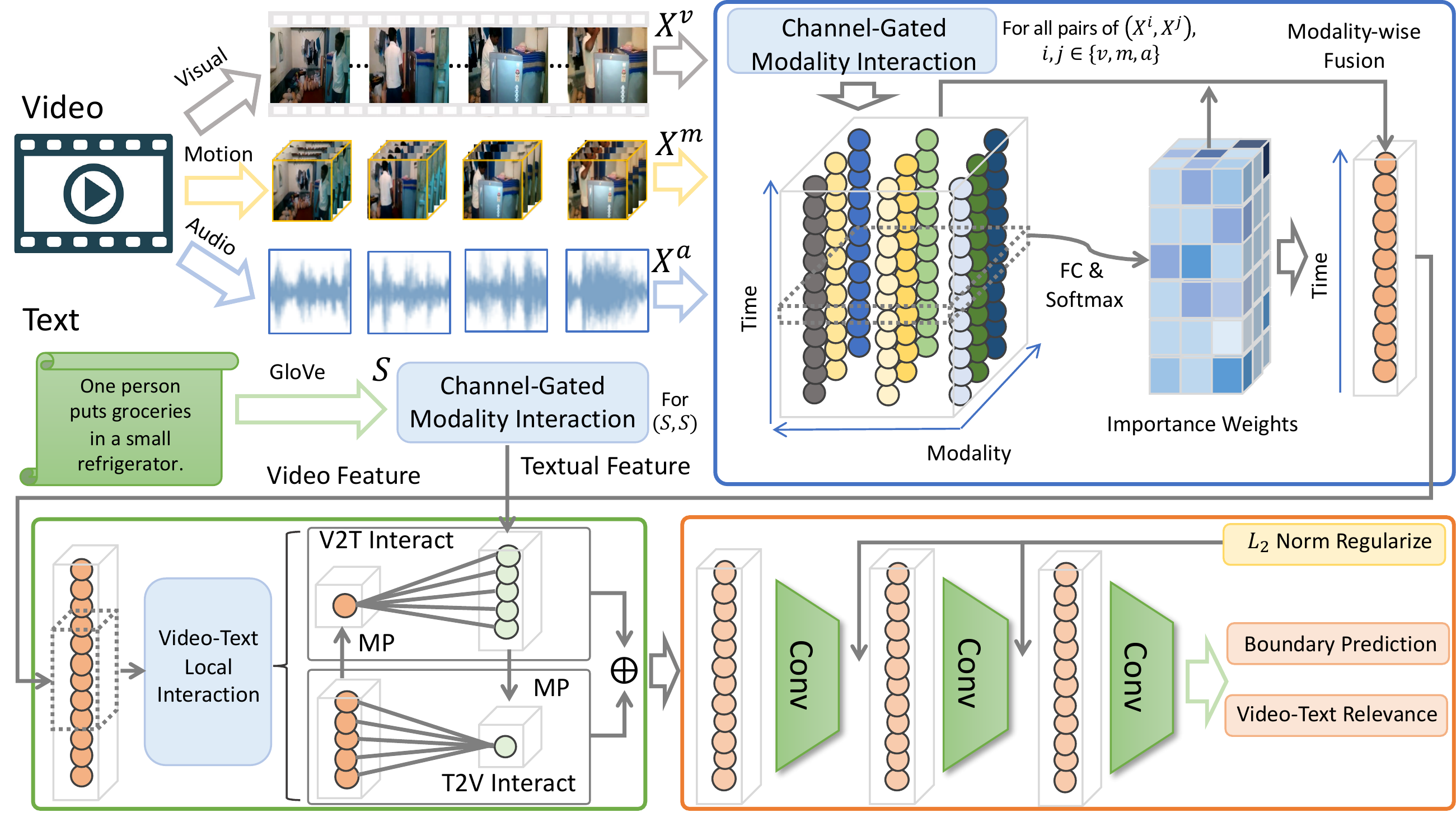}
	\caption{The framework of our approach. The multimodal features from a video are processed with Channel-Gated Modality Interaction (see Fig.~\ref{fig:interaction}) for each pair of modalities, and a weighted modality-wise fusion is then executed to obtain an aggregated video feature(\textcolor[rgb]{0.23,0.46,0.76}{Blue box}). \textbf{Note that this feature can also be used for video captioning, but the two tasks are not jointly trained}. For temporal sentence localization, the word embedding features also interact with themselves to exploit intra-sentence information, resulting in a textual feature. The video and textual features then interact locally at each temporal position (\textcolor[rgb]{0.44,0.66,0.30}{Green box}), and the resulting feature is fed to a light-weight convolutional network with layer-wise norm regularization to produce predictions (\textcolor[rgb]{0.93,0.48,0.21}{Orange box}). Each colored circle represents a feature vector.}
	\label{fig:framework}

\end{figure*}

\subsection{Overview}

We first give an overview of our approach. 
As shown in Fig.~\ref{fig:framework}, multimodal features are first extracted from a given video and then fed to a video modality interaction module, where a Channel-Gated Modality Interaction is performed for all pairs of modalities to exploit intra- and inter-modality information. The interaction results are tiled into a high-dimensional tensor and we then use a simple fully-connected network to efficiently compute the importance weights to transform this tensor into a feature sequence. This process to model pairwise modality interaction is abbreviated as PMI.

For sentence localization, the text features are also processed with modality interaction to exploit its intra-modality information. Then video and textual features are locally interacted in order to capture the complex association between these two modalities at each temporal location. Finally, a light-weight convolutional network is applied as the localization head to process the feature sequence and output the video-text relevance score and boundary prediction. 

For video captioning, since the focus of this paper is to fully exploit multimodal information, we do not adopt a sophisticated decoder architecture and only use a two-layer LSTM with temporal attention on top of the video modality interaction. However, due to the superiority of PMI, state-of-the-art performances are still achieved. Note that video modality interaction can be used in either a sentence localization model or an event captioning model, but the models are trained separately. 

\subsection{Video Modality Interaction}
Given an input video $\bm{V}=\{\bm{f}_i\}_{i=1}^{F}$, where $\bm{f}_i$ is the $i$-th frame, multimodal features can be extracted using off-the-shelf deep neural networks. 
In this paper, three apparent modalities in videos are adopted, which are visual modality, motion modality, and audio modality. Given features from these modalities, a sequence of features can be learned to represent the latent semantic modality\footnote{For fair comparison, we do not include this modality when comparing with state-of-the-art methods, but will demonstrate some qualitative results with the latent semantic modality. The corresponding learning method is placed in the Supplementary Material.}. The corresponding feature sequences from the above modalities are denoted by 
$\bm{X}^v=\{\bm{x}^v_n\}_{n=1}^N$, 
$\bm{X}^m=\{\bm{x}^m_n\}_{n=1}^N$, 
$\bm{X}^a=\{\bm{x}^a_n\}_{n=1}^N$, and
$\bm{X}^l=\{\bm{x}^l_n\}_{n=1}^N$,
respectively. 
The dimensionalities of the feature vectors in each modality are denoted as $d_v$, $d_m$, $d_a$, and $d_l$, respectively. 

\begin{figure*}[t]
	\centering \includegraphics[width=.95\textwidth]{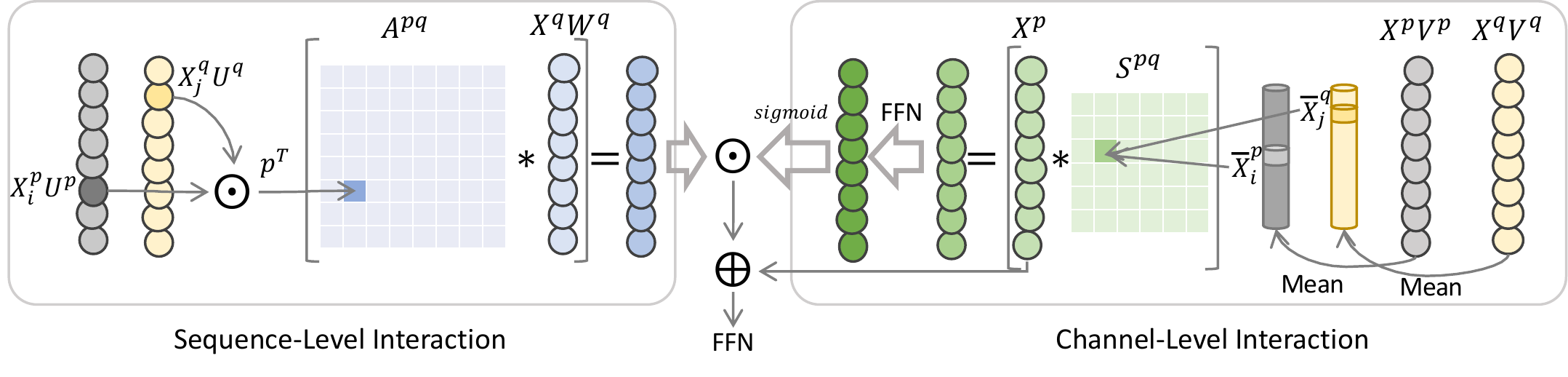}
	\caption{Overview of Channel-Gated Modality Interaction. The Channel-Level Interaction results are used as a gating variable to modulate Sequence-Level Interaction results. Details are illustrated below in Eqs~\eqref{eq1}-\eqref{eq6}.}
	\label{fig:interaction}

\end{figure*}

We propose to explicitly model modality interaction between a pair of feature sequences, denoted by $\bm{X}^p$ and $\bm{X}^q$, where $p \in \{a, m, v, l\}$ and $q \in \{a, m, v, l\}$. 
Note that $p$ and $q$ can be the same modality, and in that case, the interaction exploits intra-modality information. 
As shown in Fig.~\ref{fig:interaction}, the interaction can be formulated as
\begin{equation} \label{eq1}
\begin{split}
\mathrm{INT}(\bm{X}^p,\bm{X}^q)=\mathrm{FFN}\big(\mathrm{BA}(\bm{X}^p,\bm{X}^q) \odot \mathrm{CG}(\bm{X}^p,\bm{X}^q) \oplus \bm{X}^p \big),
\end{split}
\end{equation}
where $\mathrm{BA}(\cdot)$ is the bilinear attention model that performs sequence-level modality interaction, $\mathrm{CG}(\cdot)$ is a channel gating mechanism based on the channel-level interaction and is used to modulate the sequence-level interaction output, a residual connection is introduced with $\oplus \bm{X}^p$, and $\mathrm{FFN}(\cdot)$ is a position-wise feed-forward network that projects its input into a lower dimension\footnote{Details about this FFN can be found in the Supplementary Material.}.

\textbf{Sequence-Level Interaction} We use a low-rank bilinear model to consider the interaction between each pair of elements in feature sequences $\bm{X}^p$ and $\bm{X}^q$:
\begin{equation} \label{eq2}
\begin{split}
\bm{A}_{ij}^{pq} &= \bm{p}^T \big( \rho(\bm{X}^p_{i}\bm{U}^p) \odot \rho(\bm{X}^q_{j}\bm{U}^q) \big),\quad
\bm{\mathcal{A}}_{ij}^{pq} =  \mathrm{Softmax}_j(\bm{A}_{ij}^{pq}),
\end{split}
\end{equation}
where $\bm{X}^p_{i}$ is the $i$-th element of $\bm{X}^p$,
$\bm{X}^q_{j}$ is the $j$-th element of $\bm{X}^q$,
and $\bm{U}^p \in \mathbb{R}^{d_p \times d}$ and $\bm{U}^q \in \mathbb{R}^{d_q \times d}$ are low-rank projection matrices ($d<min(d_p,d_q)$). 
$\odot$ denotes element-wise multiplication (Hadamard product), and $\rho$ denotes ReLU non-linearity. $\bm{p}\in \mathbb{R}^{d}$ projects the element interaction into a scalar, so that $\bm{A}^{pq}\in\mathbb{R}^{N \times N}$ can be normalized into a bilinear attention map by applying column-wise softmax.
Then the output of the bilinear model is 
\begin{equation} \label{eq3}
\begin{split}
\mathrm{BA}(\bm{X}^p,\bm{X}^q)= \bm{\mathcal{A}}^{pq}(\bm{X}^q\bm{W}^q).
\end{split}
\end{equation}
In the matrix multiplication of $\bm{\mathcal{A}}^{pq}$ and $\bm{X}^q\bm{W}^q$, a relative position embedding~\cite{Shaw2018} is injected to make the sequence-level interaction to be position-aware.

\textbf{Channel-Level Interaction} In order to modulate the sequence-level interaction result, we devise a gate function based on fine-grained channel-level interaction.
We first obtain a channel representation of $\bm{X}^p$ and $\bm{X}^p$ as 
\begin{equation} \label{eq4}
\begin{split}
\overline{\bm{X}}^p=\mathrm{Mean}_n(\bm{X}^p\bm{V}^p),\quad  \overline{\bm{X}}^q=\mathrm{Mean}_n(\bm{X}^q\bm{V}^q),
\end{split}
\end{equation}
where $\mathrm{Mean}(\cdot)$ is sequence-wise mean-pooling, and $\bm{V}^p,\bm{V}^q$ are used to project $\bm{X}^p$ and $\bm{X}^p$ to lower dimension for efficient processing.
Similarly, we also compute a channel-to-channel attention map
\begin{equation} \label{eq5}
\begin{split}
\bm{S}_{ij}^{pq}=f_{chn}\big(\overline{\bm{X}}^p_i,\overline{\bm{X}}^q_j\big),\quad 
\bm{\mathcal{S}}_{ij}^{pq} =  \mathrm{Softmax}_i(\bm{S}_{ij}^{pq}),
\end{split}
\end{equation}
where $f_{chn}(\cdot)$ is a function for computing channel-level interaction. Since each element in $\overline{\bm{X}}^p$ and $\overline{\bm{X}}^q$ is a scalar, we simply use $f_{chn}(a,b)=-(a-b)^2$. Then the output of the gate function is
\begin{equation} \label{eq6}
\begin{split}
\mathrm{CG}(\bm{X}^p,\bm{X}^q)= \sigma\big(\mathrm{FFN}(\bm{X}^p\bm{\mathcal{S}}^{pq})\big),
\end{split}
\end{equation}
where $\sigma$ is the Sigmoid function, so the output has values in $[0,1]$.

\textbf{Modality-Wise Fusion.} Given $M$ modalities, there will be $M^2$ pairs of interacting modalities, 
and they are tiled as a high-dimensional tensor $\bm{X}^{MI}\in \mathbb{R}^{N\times M^2\times d}$.
The information in $\bm{X}^{MI}$ needs to be further aggregated before feeding it to target tasks. 
Simple concatenation or pooling can achieve this purpose. 
Here, we consider the importance of each interacting result by using a position-wise fully-connected layer to predict importance weights:
\begin{equation} \label{eq7}
\begin{split}
\bm{e}_{n} &= \bm{X}^{MI}_n \bm{W}^a_n+\bm{b}^a_n,\quad 
\bm{\alpha}_{n} = \mathrm{Softmax}_m(\bm{e}_{n}),\\
\bm{\widehat{X}}_n&=\sum\nolimits_{m=1}^{M^2} \bm{\alpha}_{nm}\bm{X}^{MI}_{nm}.
\end{split}
\end{equation}
Finally, the fusion result $\bm{\widehat{X}}\in \mathbb{R}^{N\times d}$ is the modality-interacted representation of a video and is ready to be used in target tasks.

\subsection{Sentence Localization}
The sentence is represented as a sequence of word-embedding vectors $Y=\{\bm{w}_l\}_{l=1}^{L}$, which is also processed with the CGMI to exploit its intra-modality information, yielding a textual feature $\widehat{\bm{Y}}$. 
For sentence localization, it is crucial to capture the complex association between the video and textual modalities at each temporal location, and then predict each location's relevance to the sentence.

\textbf{Video-Text Local Interaction} Based on the above intuition, we propose Video-Text Local Interaction. 
For each temporal location $t\in [1,N]$ of $\bm{\widehat{X}}$, 
a local window $\bm{\widetilde{X}}=\{\bm{\widehat{X}}_n\}_{n=t-w}^{t+w}$ is extracted to interact with the textual feature $\widehat{\bm{Y}}$. 
As shown in Fig.~\ref{fig:framework}, the local video-to-text interaction is modeled as 
\begin{equation} \label{eq8}
\begin{split}
\bm{Z}_t^{xy}&=\mathrm{BA}(\mathrm{Mean}(\bm{\widetilde{X}}),\widehat{\bm{Y}}),\quad 
\bm{\widehat{Z}}_t^{xy}=\mathrm{MM}(\bm{Z}_t^{xy}, \mathrm{Mean}(\bm{\widetilde{X}})).
\end{split}
\end{equation}
Here instead of gating, we use a more efficient multimodal processing unit $\mathrm{MM}(a,b)=\bm{W}^T[a||b||a\odot b||a\oplus b]$ to encourage further interaction of both modalities. Likewise, text-to-video interaction $\bm{\widehat{Z}}_t^{yx}$ is computed given $\bm{\widetilde{X}}$ and $\mathrm{Mean}(\widehat{\bm{Y}})$, and then fused with the video-to-text interaction result
\begin{equation} \label{eq9}
\begin{split}
\bm{Z}_t&=\bm{\widehat{Z}}_t^{xy} \oplus \bm{\widehat{Z}}_t^{yx}.
\end{split}
\end{equation}

\textbf{Localization Head} We apply a light-weight convolutional network upon the video-text interacted sequence $\bm{Z}$ to produce predictions. Each layer can be formulated as
\begin{equation} \label{eq10}
\begin{split}
\bm{C}^k=Conv(\bm{C}^{k-1}||\mathrm{Mean}(\widehat{\bm{Y}})),
\end{split}
\end{equation}
where $k=1,..,K$, and $\bm{C}_0=\bm{Z}$. We apply Instance Normalization~\cite{Ulyanov2016} and LeakyReLU~\cite{Xu2015a} activation to each layer's output.
Since we are computing the video-text relevance in a layer-wise fashion, we impose an $\ell_2$ norm regularization on each layer's output to obtain a more robust feature
\begin{equation} \label{eq11}
\begin{split}
\mathrm{Loss}_{norm}=\sum_{n=1}^{N}(||\bm{C}_n^k||_2-\beta_k)^2,
\end{split}
\end{equation}
where $||\cdot||$ is the $\ell_2$ norm of a vector. The $K$-th layer output $\bm{C}^K$ has 1 output channel, which is normalized using Softmax, representing the Video-Text Relevance $\bm{r}\in [0,1]^{N}$. Then a fully connected layer with two output units is applied to $\bm{r}$ to produce a boundary prediction $\bm{b}\in \mathbb{R}^{2}$.
The loss for the predictions is
\begin{equation} \label{eq12}
\begin{split}
\mathrm{Loss}_{pred}=\mathrm{Huber}(\bm{b}-\bm{\hat{b}})-\lambda_r \frac{\sum_{n} \bm{\hat{r}}_n\log(\bm{r}_n)}{\sum_{n} \bm{\hat{r}}_n},
\end{split}
\end{equation}
where $\bm{\hat{b}}$ is the ground-truth temporal boundary, $\mathrm{Huber}(\cdot)$ is the Huber loss function, and $\bm{\hat{r}}_n=1$ if $n$ is in the ground-truth temporal region, otherwise $\bm{\hat{r}}_n=0$. The overall loss is 
\begin{equation} \label{eq13}
\begin{split}
\mathrm{Loss}_{loc}=\mathrm{Loss}_{pred}+\lambda_n\mathrm{Loss}_{norm},
\end{split}
\end{equation}
where $\lambda_n,\lambda_r$ are constant weights used to balance the loss terms.

\subsection{Event Captioning}

After the video modality interaction result is obtained, we use a standard bi-directional LSTM for encoding and a two-layer LSTM network with temporal attention~\cite{Yao2015} to generate sentences as in previous works~\cite{Wang2019b,Chen2019a,Xu2017}. 
The sentence generation is done in a word-by-word fashion. 
At every time step, a set of temporal attention weights is computed based on the LSTM hidden states and video features, which is then used to weighted-sum the video features into a single vector.
This dynamic feature vector is fed to the LSTM with the previously generated word to predict the next word\footnote{Due to the space limit and that caption decoder is not the focus of this work, we omit formal descriptions here. We also move some experiments and analysis below to the Supplementary Material.}.
We would like to emphasize again that video modality interaction can be used as a basic video feature encoding technique for either sentence localization or event captioning, but we do not perform multi-task training for these two.

\section{Experiments}
In this section, we provide experimental analysis of our model design and present comparisons with the state-of-the-art methods on both temporal sentence localization and video captioning.

\subsection{Experimental Settings}

\textbf{MSVD Dataset}~\cite{Chen2011}.
MSVD is a well-known video captioning dataset with 1,970 videos. 
The average length is 9.6 seconds, and each video has around 40 sentence annotations on average. 
We adopt the same common dataset split as in prior works~\cite{Yao2015,Xu2017,Baraldi2017}. 
Thus, we have $1,200$ / $100$ / $670$ videos for training, validation, and testing, respectively.

\textbf{MSR-VTT Dataset}~\cite{Xu2016}.
MSR-VTT is a large-scale video captioning dataset with $10,000$ videos.
The standard split~\cite{Xu2016} for this dataset was provided. 
Hence, we use $6,513$ / $497$ / $2,990$ videos for training, validation, and testing, respectively, in our experiments. In this dataset,  each video is associated with 20 sentence annotations and is of lenght $14.9$ seconds on average. 

\textbf{ActivityNet Captions Dataset}~\cite{Krishna2017}(ANet-Cap).
ANet-Cap is built on the ActivityNet dataset~\cite{Heilbron2015} with $19,994$ untrimmed videos ($153$ seconds on average). 
The standard split is $10,009$ / $4,917$ / $5,068$ videos  for training, validation, and testing, respectively.
There are $3.74$~\textit{temporally localized} sentences per video on average. 
Since the testing set is not publicly available, we evaluate our method on the validation set as previous works~\cite{Wang2020,Xu2019}.

\textbf{Charades-STA Dataset}~\cite{Gao2017}.
Charades-STA is built on $6,672$ videos from the Charades~\cite{Sigurdsson2016} dataset. 
The average duration of the videos is $29.8$ seconds. 
There are $16,128$ \textit{temporally localized} sentence annotations,  which give $2.42$ sentences per video.
The training and testing sets contain $12,408$ and $3,720$ annotations, respectively. 

We evaluate the captioning performance of our method on MSVD and MSR-VTT with commonly used metrics, \textit{i.e.,} BLEU~\cite{Papineni2002}, METEOR~\cite{Denkowski2014}, and CIDEr~\cite{Vedantam2015}. 
ANet-Cap and Charades-STA are used to evaluate sentence localization performance.
We adopt the same evaluation metric used by previous works~\cite{Gao2017},
which computes ``Recall@1,IoU=m'' (denoted by $r(m, s_i)$),
meaning the percentage of the top-1 results 
having IoU larger than $m$ with the annotated segment of a sentence $s_i$.
The overall performance on a dataset of $N$ sentences 
is the average score of all the sentences $\frac{1}{N} \sum_{i=1}^{N}r(m, s_i)$. 

\textbf{Implementation Details.}
The sentences in all datasets are converted to lowercase and then tokenized. For the captioning task, randomly-initialized word embedding vectors of dimension 512 are used, which are then jointly fine-tuned with the model. 
For the sentence localization task, we employ the GloVe~\cite{Pennington2014} word embedding as previous works. 
We use Inception-ResNet v2~\cite{Szegedy2017} and C3D~\cite{Tran2015} to extract visual and motion features. 
For the audio features, we employ the MFCC (Mel-Frequency Cepstral Coefficients) on the captioning task and SoundNet~\cite{Aytar2016} on the sentence localization task.
We temporally subsample the feature sequences to length 32 for event captioning, and 128 for sentence localization. 
The bilinear attention adopts 8 attention heads, and the loss weights $\lambda_r$ and $\lambda_n$ are set to 5 and 0.001, respectively.
In all of our experiments, the batch size is set to 32 and  the Adam optimizer with learning rate 0.0001 is used to train our model. 

\begin{table}[t]
	\scriptsize
	\centering
	\caption{Performance comparison of video modality interaction strategies on MSVD.}
	\begin{tabular}{|c|l | c c c|} 
		\hline
		\#&Method & B@4 & M & C\\
		\hline
		0&Concat w/o Interact (Baseline) & 45.28 & 31.60 & 62.57\\
		1&Concat + Interact & 46.24 & 32.03 &66.10\\
		2&Pairwise Interact + Concat Fusion& 47.86 & 33.73 & 75.30\\
		3&Pairwise Interact + Sum Fusion & 51.37 &34.01& 78.42\\
		4&Pairwise Interact + Weighted Fusion (ours) &54.68 & 36.40 &95.17 \\
		\hline
		5&Intra-modality Interactions only &  49.92 & 34.76& 88.46 \\
		6&Inter-modality Interactions only &  47.30& 32.72 & 70.20\\
		7&(Intra+Inter)-modality (ours) & 54.68 & 36.40 &95.17 \\
		\hline
	\end{tabular}

	\label{table:1}

\end{table}

\begin{table}[t]
	\scriptsize
	\centering
	\caption{Performances (\%) of different localizer settings on the Charades-STA dataset.}
	\begin{tabular}{|l|c | c | c |c c c|} 
		\hline
		\#&PMI & VTLI & $\ell_2$-Norm & IoU=0.3 & IoU=0.5 & IoU=0.7\\
		\hline
		0&\xmark &\xmark &\xmark & 51.46 & 35.34 & 15.81\\
		1&\cmark &\xmark &\xmark & 53.22 & 37.05 & 17.36\\
		2&\cmark &\cmark &\xmark & 54.37 & 38.42 & 18.63\\
		3&\cmark &\cmark &\cmark & 55.48 & 39.73 & 19.27\\
		\hline
	\end{tabular}
	\label{table:2}

\end{table}

\subsection{Ablation Studies}
Firstly, we perform extensive experiments to validate the design choices in our approach. 
We study the effect of different modality interaction strategies on the MSVD dataset, 
and the effects of sentence localizer components on the Charades-STA dataset. 
All experiments use Inception-ResNet v2 and C3D features. 

On the MSVD dataset, we design 8 different variants and their performances are summarized in Table~\ref{table:1}. 
In variant 0, which is a baseline, multimodal features are concatenated and directly fed to the caption decoder. 
Variant 1 treats the concatenated features as one modality and performs intra-modality interaction.
In variants 2-4, PMI is performed and different fusion strategies are adopted. 
In variants 5-7, we study the ablation of intra- and inter-modality interactions.

\textbf{Why pairwise?} 
We perform modality interaction in a pairwise fashion in our model, and this is the main distinctive difference from existing methods~\cite{Song2019,Ye2019}, which employ feature concatenation. 
As shown in Table~\ref{table:1}, 
while concatenating all modalities into one and performing intra-modality interaction
can gain performance improvements over the baseline (\#1 vs. \#0),
concatenating after pairwise interaction
has a more significant advantage (\#2 vs. \#1). 
We also compare the effects of different aggregation strategies 
after pairwise interaction (\#2-4), 
and weighted fusion (in PMI) yields the best result with a clear margin,
which also indicates that the interactions between different modality pairs
produce unique information of different importances.

\textbf{Effect of inter-modality complementarity.}
We then inspect the intra- and inter-modality interactions separately.
Table~\ref{table:1} (\#5-7) shows that intra-modality interaction 
can already effectively exploit information in each modality compared to the baseline. 
Inter-modality complementarity alone is not sufficient for captioning,
but it can be combined with intra-modality information 
to obtain a further performance boost, 
which again validates our design of pairwise interaction.


\begin{table}[t]
	\scriptsize
	\centering
	\caption{Video captioning performances of our proposed PMI and other state-of-the-art multimodal fusion methods on the MSVD dataset. Meanings of features can be found in Table~\ref{table:5}.}
	\begin{tabular}{|c | c | c c c|} 
		\hline
		Method & Features & B@4 & M & C\\
		\hline
		AF~\cite{Hori2017} & V+C & 52.4 & 32.0 & 68.8\\
		TDDF~\cite{Zhang2017}& V+C & 45.8 & 33.3 & 73.0 \\
		MA-LSTM~\cite{Xu2017} & G+C & 52.3 & 33.6 & 70.4 \\
		MFATT~\cite{Long2018} & R152+C & 50.8 & 33.2 &69.4\\
		GRU-EVE~\cite{Aafaq2019} & IRV2+C & 47.9 & 35.0 & 78.1 \\
		XGating~\cite{Wang2019b} & IRV2+I3D & 52.5 & 34.1 & 88.7 \\
		HOCA~\cite{Jin2019}	&IRV2+I3D & 52.9 & 35.5 & 86.1 \\
		\hline
		PMI-CAP & V+C & 49.74 & 33.59 & 77.11 \\
		PMI-CAP & G+C & 51.55 & 34.64 & 74.51 \\
		PMI-CAP & R152+C & 52.07 & 34.34 & 77.35 \\
		PMI-CAP & IRV2+C & 54.68 & 36.40 &95.17\\
		PMI-CAP & IRV2+I3D & 55.76 & 36.63 & 95.68  \\ 
		\hline
	\end{tabular}

	\label{table:4}

\end{table}

\begin{table}[t]
	\scriptsize
	\centering
	\caption{Performances of our proposed model and other state-of-the-art methods on the MSVD and MSR-VTT datasets. R*, G, V, C, IV4, R3D, IRV2, Obj, and A mean ResNet, GoogLeNet, VGGNet, C3D, Inception-V4, 3D ResNeXt, Inception-ResNet v2, Object features, and audio features, respectively. Note that audio track is only available on MSR-VTT, and for fair comparison, we use the MFCC audio representation as~\cite{Chen2019a,Chen2017}. Please refer to the original papers for the detailed feature extraction settings.}
	\begin{tabular}{|c |c| c c c| c| c c c |} 
		\hline
		Dataset & \multicolumn{4}{c|}{MSVD} & \multicolumn{4}{c|}{MSR-VTT} \\
		\hline
		Method & Features & B@4 & M & C&Features & B@4 & M & C\\
		\hline
		STAT~\cite{Tu2017}		& G+C+Obj & 51.1 &32.7& 67.5 & G+C+Obj &37.4& 26.6 &41.5 \\
		M$^3$~\cite{Wang2018} 	& V+C &51.78 & 32.49&- 	& V+C & 38.13 & 26.58 & -\\
		DenseLSTM~\cite{Zhu2019}& V+C & 50.4 &32.9& 72.6 & V+C & 38.1 &26.6& 42.8 \\
		PickNet~\cite{Chen2018a}& R152& 52.3 & 33.3 & 76.5  & R152  & 41.3& 27.7& 44.1 \\
		hLSTMat~\cite{Song2017}	& R152& 53.0 &33.6&73.8 & R152&38.3 &26.3&- \\
		VRE~\cite{Shi2019} 		& R152&51.7 &34.3& 86.7 & R152+A&43.2 & 28.0 &48.3\\
		MARN~\cite{Pei2019} 	& R101+R3D &  48.6 & 35.1 &\underline{92.2} & R101+R3D & 40.4 &28.1 &47.1 \\
		OA-BTG~\cite{Zhang2019}	& R200+Obj &\textbf{56.9}&\underline{36.2}&90.6 & R200+Obj & 41.4 & 28.2&46.9 \\
		RecNet~\cite{Wang2018a} & IV4 &52.3&34.1&80.3 & IV4 & 39.1&26.6 & 42.7 \\
		XGating~\cite{Wang2019b}& IRV2+I3D & 52.5 & 34.1 & 88.7 & IRV2+I3D & 42.0 & 28.1 & 49.0 \\
		MM-TGM~\cite{Chen2017} 	& IRV2+C &48.76&34.36&80.45 & IRV2+C+A&\underline{44.33}&\underline{29.37}&49.26 \\
		GRU-EVE~\cite{Aafaq2019}& IRV2+C &47.9&35.0&78.1 & IRV2+C & 38.3 & 28.4 &48.1 \\
		MGSA~\cite{Chen2019a} 	& IRV2+C &53.4&35.0&86.7 & IRV2+C+A& \textbf{45.4}&28.6&\underline{50.1} \\
		\hline
		
		PMI-CAP	&IRV2+C&\underline{54.68}&\textbf{36.40}&\textbf{95.17}	&IRV2+C	& 42.17 & 28.79 & 49.45\\
		PMI-CAP &-&-&-&-&IRV2+C+A& 43.96 & \textbf{29.56} & \textbf{50.66}\\
		\hline
	\end{tabular}

	\label{table:5}

\end{table}

\textbf{Effect of sentence localizer components.}
The PMI, video-text local interaction (VTLI), and $\ell_2$-norm regularization are the key components of the sentence localization model. As can be observed from Table~\ref{table:2}, incorporating each component consistently leads to a performance boost.


\subsection{Comparison with State-of-the-Art Methods}

\textbf{Results on the Video Event Captioning Task.}
We abbreviate our approach as PMI-CAP for video captioning. To demonstrate the superiority of our proposed pairwise modality interaction, we first compare our method with state-of-the-art methods that focus on the fusion of multimodal features for video captioning. For fair comparison, we use the same set of features as each compared method.
As shown in Table~\ref{table:4}, our PMI-CAP has outperformed all the compared methods when using the same features. 
The improvement in the CIDEr metric is especially significant, which is 10.8\% on average. 
This shows that our pairwise modality interaction can really utilize multimodal features more effectively. 

Table~\ref{table:5} shows the performance comparison on the MSVD and MSR-VTT datasets.
We adopt the set of features commonly used by recent state-of-the-art methods~\cite{Wang2019b,Aafaq2019,Chen2019a}, 
which are Inception-ResNet v2 and C3D for visual and motion modalities, respectively.
Among the competitive methods, OA-BTG~\cite{Zhang2019} utilizes object-level information from an external detector, and MARN~\cite{Pei2019} uses a more advanced 3D CNN to extract motion features.
We do not exploit spatial information like MGSA~\cite{Chen2019a} and VRE~\cite{Shi2019},
or use a sophisticated decoder as hLSTMat~\cite{Song2017} and MM-TGM~\cite{Chen2017}, 
while we emphasize that PMI may be used along with most of these methods. 
Overall, our PMI-CAP achieves state-of-the-art performances on both MSVD and MSR-VTT.

\begin{table}[t]
	\scriptsize
	\centering
	\caption{Performances (\%) of our proposed model and other state-of-the-art methods on the Charades-STA dataset. * means our implementation.}
	\begin{tabular}{|c|c c c|} 
		
		\hline
		Method& IoU=0.3 & IoU=0.5 & IoU=0.7 \\
		\hline
		Random & 14.16 & 6.05 & 1.59 \\
		VSA-RNN~\cite{Karpathy2015} &-&10.50&4.32\\
		VSA-STV~\cite{Karpathy2015} &-&16.91&5.81\\
		MCN~\cite{Hendricks2017} &32.59& 11.67& 2.63 \\
		ACRN~\cite{Liu2018} &38.06& 20.26& 7.64 \\
		ROLE~\cite{Liu2018a} &37.68&21.74&7.82 \\
		SLTA~\cite{Jiang2019} &38.96&22.81&8.25 \\
		CTRL~\cite{Gao2017} &-&23.63& 8.89\\
		VAL~\cite{Song2018} &-&23.12&9.16\\
		ACL~\cite{Ge2019} &-&30.48&12.20\\
		SAP~\cite{Chen2019} &-&27.42&13.36\\
		SM-RL~\cite{Wang2019} &-&24.36&11.17\\
		QSPN~\cite{Xu2019} &54.7&35.6&15.8\\
		ABLR*~\cite{Yuan2019} &51.55 &35.43 &15.05 \\
		TripNet~\cite{Hahn2019} &51.33&36.61&14.50\\
		CBP~\cite{Wang2020} &-&36.80&18.87\\
		\hline
		PMI-LOC (C) & \textbf{55.48}& \textbf{39.73}& \textbf{19.27} \\
		\hline
		PMI-LOC (C+IRV2) & 56.84 & 41.29 & 20.11   \\
		PMI-LOC (C+IRV2+A) & 58.08 & 42.63 & 21.32  \\
		\hline
	\end{tabular}

	\label{table:7}

\end{table}

\begin{table}[t]
	\scriptsize
	\centering
	\caption{Performances (\%) of our proposed model and other state-of-the-art methods on the ActivityNet Captions dataset.}
	\begin{tabular}{|c|c c c|} 
		\hline
		Method& IoU=0.3 & IoU=0.5 & IoU=0.7 \\
		\hline
		Random &12.46&6.37&2.23\\
		QSPN~\cite{Xu2019} &45.3&27.7&13.6\\
		TGN~\cite{Chen2018} &43.81&27.93&-\\
		ABLR~\cite{Yuan2019}&55.67&36.79&-\\
		TripNet~\cite{Hahn2019} &48.42&32.19&13.93\\
		CBP~\cite{Wang2020} &54.30&35.76&17.80\\
		\hline
		PMI-LOC (C) & \textbf{59.69} & \textbf{38.28} & \textbf{17.83} \\
		\hline
		PMI-LOC (C+IRV2) & 60.16 & 39.16 & 18.02  \\
		PMI-LOC (C+IRV2+A) & 61.22 & 40.07 & 18.29 \\
		\hline
	\end{tabular}

	\label{table:8}

\end{table}


\textbf{Results on the Sentence Localization Task.}
As previously introduced, current state-of-the-art methods for sentence localization haven't considered this problem in a multimodal setting and only use the C3D feature.
Thus we present results with only C3D feature to fairly compare with these methods and also report performances under multimodal settings. 
Our approach is abbreviated as PMI-LOC for sentence localization.
Table~\ref{table:7} shows results on the widely-used Charades-STA dataset. 
Our PMI-LOC outperforms all compared methods in all metrics. 
Further experiments with multimodal features show even higher localization accuracies, which verify the effectiveness of our modality interaction method. 
As shown in Table~\ref{table:8}, on the large-scale ActivityNet Captions dataset, our method also achieves state-of-the-art performances. 

\begin{figure*}[t]
	\centering \includegraphics[width=.85\textwidth]{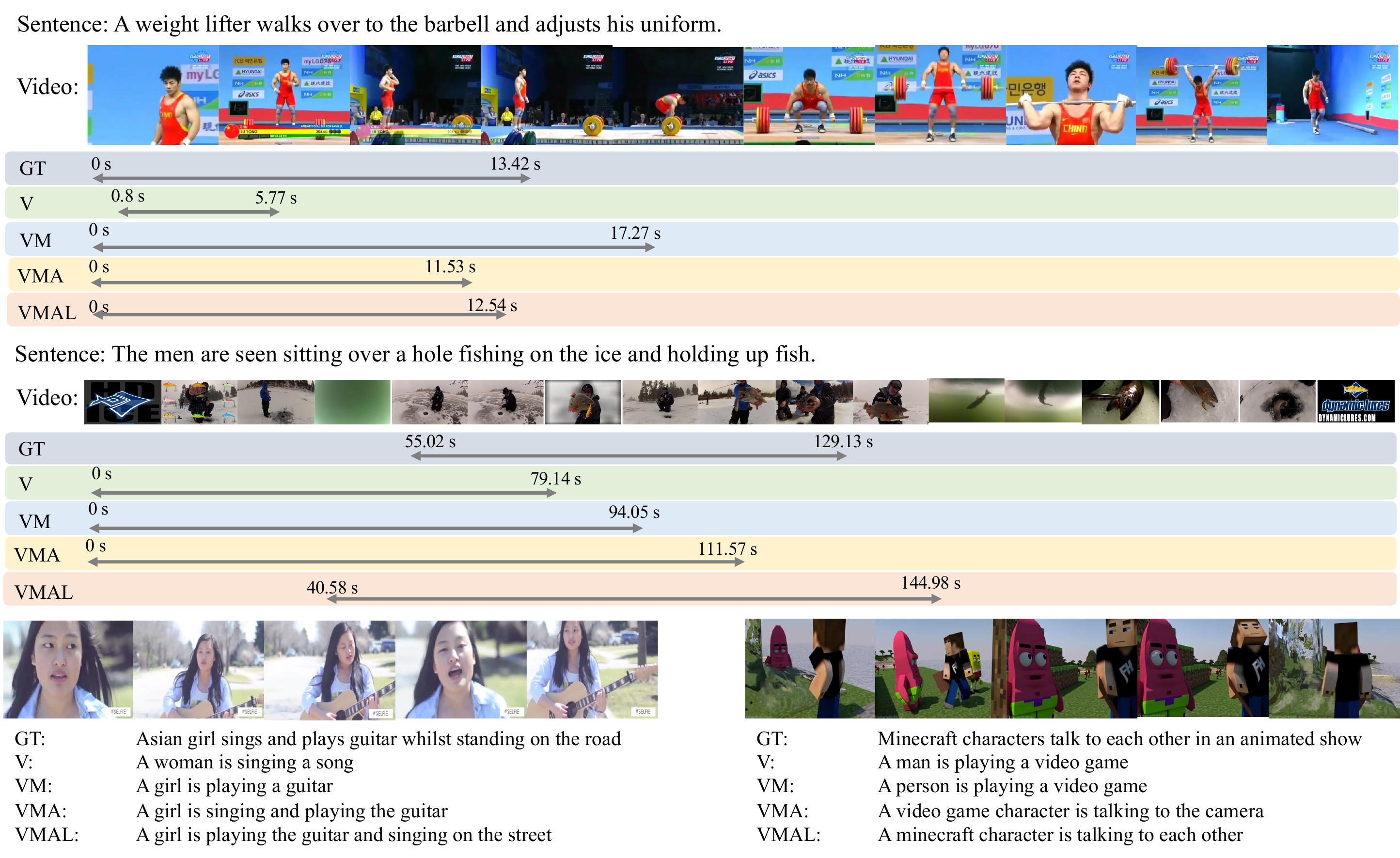}
	\caption{Qualitative results of temporal sentence localization and event captioning. The results are generated using our model but with different combinations of modalities.}
	\label{fig:qua_1}
\end{figure*}

\begin{figure*}[t]
	\centering \includegraphics[width=.80\textwidth]{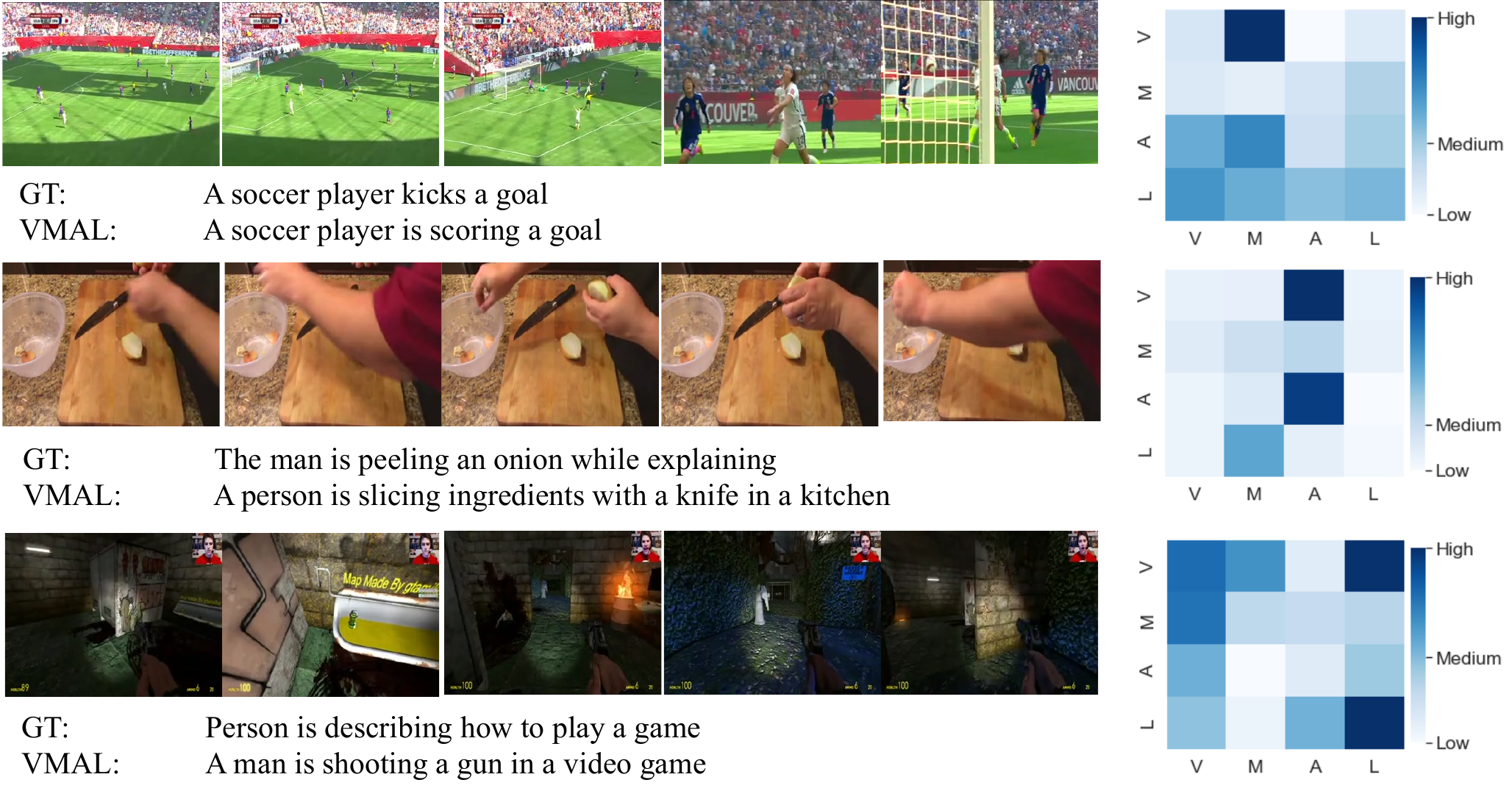}
	\caption{Qualitative results of video event captioning with visualization of the modality importance weights.}
	\label{fig:qua_2}
\end{figure*}

\begin{figure*}[t]
	\centering \includegraphics[width=.90\textwidth]{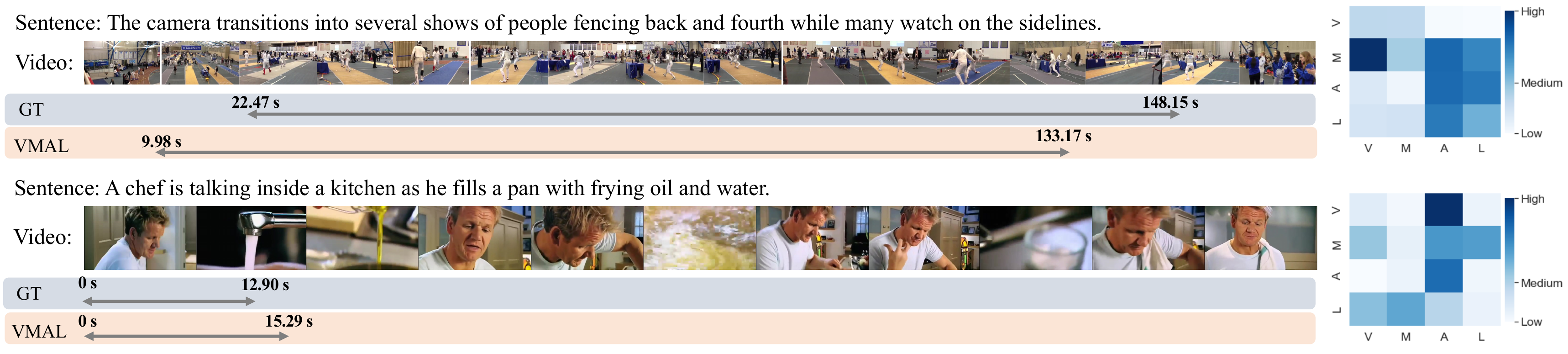}
	\caption{Qualitative results of temporal sentence localization with visualization of the modality importance weights.}
	\label{fig:qua_3}
\end{figure*}

\subsection{Qualitative Results}
We show some qualitative results in Figures~\ref{fig:qua_1}, ~\ref{fig:qua_2}, and~\ref{fig:qua_3} to demonstrate the effectiveness of our modality interaction method and how it provides expainability to the final prediction of the target tasks. Note that in addition to the visual (V), motion (M), and audio (A) modalities, we also utilize the previously mentioned latent semantics (L) modality to comprehensively explore the video content. 

Fig.~\ref{fig:qua_1} indicates that by utilizing more modalities, the model gets more complementary information through modality interaction and achieves better performance for both temporal sentence localization and event captioning. 
The event captioning examples in Fig.~\ref{fig:qua_2} show that each type of events has its modality interaction pattern. 
The sports video (top) has distinctive visual and motion patterns that are mainly captured by visual-motion modality interaction. The cooking video (middle) has unique visual cues and sounds made by kitchenware, so the important interactions are between the visual and audio modalities and within the audio modality. For the animated video (bottom),  latent semantics modality is important when the other modalities are not sufficient to capture its contents. 
Similar observations can also be made on the sentence localization examples in Fig.~\ref{fig:qua_3}.

\section{Conclusions}
In this paper, we proposed pairwise modality interaction (PMI) for tackling the temporal sentence localization and event captioning tasks, and performed fine-grained cross-modal interactions in both the sequence and channel levels to better understand video contents.
The extensive experiments on four benchmark datasets on both tasks consistently verify the effectiveness of our proposed method. 
Our future work will extend the proposed modality interaction method to cope with other video understanding tasks.

\section*{Acknowledgement}
Shaoxiang Chen is partially supported by the Tencent Elite Internship program.

\clearpage
%
%
\bibliographystyle{splncs04}
\bibliography{references}

\clearpage


\section*{Supplementary material}

# Supplementary Material for: Learning Modality Interaction for Temporal Sentence Localization and Event Captioning in Videos

Shaoxiang Chen[1][*], Wenhao Jiang[2], Wei Liu[2], and Yu-Gang Jiang[1][**]

[1] Shanghai Key Lab of Intelligent Information Processing,
School of Computer Science, Fudan University
[2] Tencent AI Lab
{sxchen13, ygj}@fudan.edu.cn, cswhjiang@gmail.com, wl2223@columbia.edu

## 1 Learning Latent Semantic Modality

Apart from the visual, motion, and audio modalities, which can be directly observed (apparent modalities), the latent semantics modality that carries high-level semantic information can be helpful for the language related tasks. We design a lightweight network to perform semantic attributes prediction using the sentence annotation provided by each dataset (either the video captioning or the sentence localization dataset). Note that this is a standalone task and the latent semantics modality is optional for our method.

The input of this network is the concatenation of all apparent modalities $\boldsymbol{X}^A = [\boldsymbol{X}^v || \boldsymbol{X}^m || \boldsymbol{X}^a]$, where $\boldsymbol{X}^A \in \mathbb{R}^{N \times (d_a + d_m + d_v)}$. We simply process $\boldsymbol{X}^A$ using bidirectional LSTMs and concatenate the hidden states of each LSTM:

$$\boldsymbol{X}^l = [\overrightarrow{\mathrm{LSTM}}(\boldsymbol{X}^A) || \overleftarrow{\mathrm{LSTM}}(\boldsymbol{X}^A)], \tag{1}$$

where $\boldsymbol{X}^l \in \mathbb{R}^{N \times 2d_{hid}}$, and $\overrightarrow{\mathrm{LSTM}}(\cdot)$ and $\overleftarrow{\mathrm{LSTM}}(\cdot)$ denote the LSTM networks that have $d_{hid}$ units and process their input sequences in the forward and backward directions, respectively. $\boldsymbol{X}^l$ is then passed through a fully-connected layer with sigmoid activation to predict semantic attribute probabilities:

$$\boldsymbol{P} = \mathrm{sigmoid}(\boldsymbol{X}^l \boldsymbol{W}_c + \boldsymbol{b}_c), \tag{2}$$

where $\boldsymbol{W}_c$ and $\boldsymbol{b}_c$ are parameters, $\boldsymbol{P} \in \mathbb{R}^{C \times N}$ collects the temporal semantic attributes, and $C$ is the vocabulary size of predefined attributes.

To train this network, we construct labels from the sentence annotations in event captioning or sentence localization datasets. We first process the training sentences of a dataset, select the most frequent $C$ words that are noun or verb, and lemmatize them to form an attribute vocabulary. Then each sentence can

---

be converted to a one-hot label $\boldsymbol{l} \in \mathbb{R}^C$ according to whether its words are in the vocabulary, where $\boldsymbol{l}_c = 1$ indicates that attribute $c$ is present in the sentence, otherwise $\boldsymbol{l}_c = 0$. The label $\boldsymbol{l}$ is broadcast to the $N$ temporal locations to compute the cross entropy loss at each location:

$$\mathcal{L}_{ce} = -\frac{1}{C} \sum\nolimits_{c=1}^{C} (\boldsymbol{l}_c \ln \boldsymbol{P}_c + (1 - \boldsymbol{l}_c) \ln(1 - \boldsymbol{P}_c)), \tag{3}$$

where $\mathcal{L}_{ce} \in \mathbb{R}^N$. For the sentence localization task, the sentence annotations are usually available for temporal segments. To unify the loss representations, we construct a temporal mask $\boldsymbol{M}^{tcp} \in [0, 1]^N$ defined as:

$$\boldsymbol{M}_i^{tcp} = \begin{cases} 1 & \text{if } i \in [s, e] \text{ and } rand(0, 1) > 0.5 \\ 0 & \text{otherwise} \end{cases} \tag{4}$$

where $[s, e]$ is the temporal segment of the sentence annotation normalized to be in $[0, 1]$. In event captioning it is safe to assume $s = 0$ and $e = 1$ since the videos are relatively short. Randomness is introduced in $\boldsymbol{M}^{tcp}$ to prevent overfitting. The final temporal semantic attributes prediction loss is computed as:

$$\mathcal{L}_{tcp} = \frac{1}{N} \mathcal{L}_{ce} \cdot \boldsymbol{M}^{tcp}, \tag{5}$$

where $\cdot$ is the dot product operator. From the above description, we can see that when the network learns to predict attributes, $\boldsymbol{X}^l$ carries rich information of latent semantics for every temporal location. Thus, it can be used to assist our target tasks through interacting with other modalities.

## 2    Feed-Forward Network (FFN)

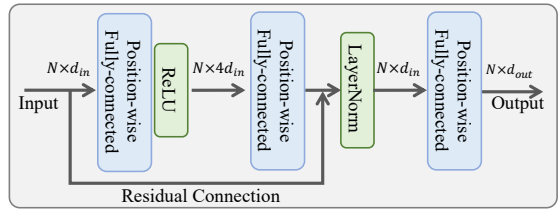

**Fig. 1.** The structure of the Feed-Forward Network (FNN).

As shown in Fig. 1, FFN is mainly composed of three position-wise fully-connected layers, each of which basically applies a fully-connected layer to each element of the input feature sequence with shared parameters. ReLU activation and layer normalization are applied to the first and second layers, respectively, and the initial input is connected to the second layer's output via a residual connection to encourage gradient flow. The output dimension $d_{out}$ is decided according to the input ($d_{out} \leq d_{in}$).

## 3   More on Motivation

Our motivation is two-fold (Note that the Equations, Tables, and Figures mentioned in this section are all in the original paper):

(1) It is intuitive that both human and AI models understand events better via a combination of different sensory modalities, but the importances of different modalities vary among videos as well as among the moments inside a video. This motivates us to fuse the modality-interacted tensor by considering both the modality-wise and sequence-wise importances (Eq. (7)).

(2) Neuroscience researches have proven that information processing in one modality can affect another, which means that there are interactions among modalities and complementary information may communicate through such interactions. This motivates us to design sequence- and channel-level interactions for each pair of modalities. In the sequence-level interaction, each element from one sequence interacts with all the elements in the other sequence through the bilinear model (Eq. (2)). This fully-connected information flow between two modalities enables better utilization of complementary information than traditional fusion strategies as shown in Table 1. It is also widely accepted that different feature channels capture different information. Thus the goal of channel-level interaction is to emphasize important channels, which is realized by gating. The gate variable is computed via a channel-to-channel attention mechanism, and sequence-wise mean-pooling (Eq. (4)) is for reducing computation. The gating power is demonstrated by the experiment below in Section 6. The improvement brought by channel-level interaction is not as significant as sequence-level interaction, but it is indeed effective.

Based on the motivation, our goal is finding a better combination of modalities via fine-grained interaction. Attention is the building block we adopted to achieve this goal, because it is easy to understand and implement (also yields a clear framework). Finally, we have proven our modality interaction to be both effective and able to provide explainability (see Figs. 4 and 5).

## 4   Computational Complexity

**Table 1.** PMI-CAP's running times on one RTX 2080Ti GPU.

| Mode | Memory | Time/batch |
|------|--------|------------|
| Train (batch size=32) | 5939MB | 0.38s |
| Infer (batch size=1) | 1441MB | 0.08s |

The major computational cost is from sequence-level interaction, which mainly consists of feature projection and bilinear modeling (Eqs. (2) and (3) in the original paper). Assume that a pair of interacting feature sequences both have dimension $b \times n \times d$, where $b$ and $n$ stand for batch size and sequence length,

**Table 2.** Performances of PMI combined with other methods of target tasks.

| Method | B@4 | M | C |
|---|---|---|---|
| Masked Transformer [3] | 47.49 | 32.43 | 77.35 |
| Masked Transformer [3]+PMI | 50.95 | 35.20 | 86.61 |

| Method | IoU=0.3 | IoU=0.5 | IoU=0.7 |
|---|---|---|---|
| ABLR [2] | 53.55 | 37.47 | 16.21 |
| ABLR [2]+PMI | 55.26 | 39.52 | 16.88 |

**Table 3.** Performance comparison on the ActivityNet Captions dataset.

| Method | B@4 | M | C |
|---|---|---|---|
| vanilla-CAP (IRV2+I3D) | 1.75 | 10.14 | 40.63 |
| PMI-CAP (IRV2+I3D) | 1.99 | 10.89 | 43.56 |
| PMI-CAP (IRV2+I3D+A) | 2.31 | 11.00 | 51.30 |
| PMI-CAP/no-channel (IRV2+I3D) | 2.00 | 10.52 | 43.06 |
| 2019 Rank-1 Intra-Event | 3.91 | 11.96 | 49.56 |

respectively. Then the computational complexity is $O(bnd^2 + bn^2d)$. For short videos, since $n \ll d$, the complexity becomes $O(bnd^2)$ (mainly batch matrix multiplication) and is efficient to run on GPU. While for very long videos like TV shows the $O(bn^2d)$ term is dominant and the computational cost would grow quadratically with video length. Nonetheless, reducing the quadratic complexity for very long videos is out of scope of this work and is left for future work. Actual running times of PMI-CAP are shown in Table 1.

## 5   Compatibility with Other Models

We also test the effectiveness of our proposed PMI when combined with other types of architectures for event captioning or sentence localization. Note that the original methods used concatenated features [3] or single feature [2] as inputs, our implementations concatenate multimodal features for both methods. Results are presented in Table 2. For video captioning, we adopt the Masked-Transformer model [3] which is essentially different from RNN-based caption decoders. We use PMI to encode the multimodal features as its input and a substantial performance improvement over feature concatenation is obtained. We combine PMI with the state-of-the-art RNN-based sentence localization method ABLR [2] by inserting our PMI module between the feature extraction and Bi-LSTM feature encoding of ABLR, and a clear performance gain is also observed.

## 6   Captioning Performances on ActivityNet Captions

We further evaluate several variants of our PMI-CAP on the ActivityNet Captions and compare them with the 2019 ActivityNet captioning challenge winner [1], which used a more diverse set of features (e.g., objects and contexts) in

addition to the three common modalities. Following the official evaluation protocol, we compare the performances of captioning ground-truth event proposals on the validation set. The results are shown in Table 3. The vanilla-CAP method removes PMI and uses feature concatenation instead. The channel-level interaction and gating are disabled in the "no-channel" setting. As can be observed from the top four rows, our proposed method is consistently effective on ActivityNet. It is notable that our method can also achieve comparable performances with the challenge winner despite using fewer features.



\end{document}